\documentclass{llncs}
\usepackage{pslatex}
\usepackage{url}
\usepackage{amsmath}

\title{Optimality Theory as a Framework \\ for Lexical Acquisition}
\titlerunning{Optimality Theory as a Framework for Lexical Acquisition}
 
\author{Thierry Poibeau
}

\institute{Laboratoire LATTICE\\
PSL*: Paris Sciences et Lettres 
\thanks{This work has received support of TransferS (laboratoire d?excellence, program ``Investissements d'avenir'' ANR-10-IDEX-0001-02 PSL* and ANR-10-LABX-0099)}  \\
1 rue Maurice Arnoux \\
92120 Montrouge France \\
\emph{thierry.poibeau@ens.fr}
}
\begin{document}

\maketitle

\begin{abstract}
This paper re-investigates a lexical acquisition system initially developed for French. We show that, interestingly, the architecture of the system reproduces and implements the main components of Optimality Theory. However, we formulate the hypothesis that some of its limitations are mainly due to a poor representation of the constraints used. Finally, we show how a better representation of the constraints used would yield better results. 
\end{abstract}

\section{Introduction}

Natural Language Processing (NLP) aims at developing techniques for processing natural language texts using computers. In order to yield accurate results, NLP requires resources containing various information (sub-categorization frames, semantic roles, selection restrictions, etc.). Unfortunately, such resources are not available for most languages and are very costly to develop manually. A recent trend of research has tried to overcome these limitations through the development of automatic acquisition methods from corpora. 

Automatic lexical acquisition is an engineering task aiming at providing compre\-hensive---even if not fully accurate---resources for NLP. As natural languages are complex, lexical acquisition needs to take into account a wide range of parameters and constraints. However, surprisingly, in the acquisition community, relatively few investigations have been done on the structure of the linguistic constraints themselves, beyond the engineering point of view (but note that this work has been extensively done for parsing, see \cite{aarts08}). 

In this paper, we want to take another look at some experiments recently done on the automatic acquisition of lexical resources from textual corpora, more specifically on French. In a way, acquisition is converse to parsing: the task consists, from a surface form, in trying to find an abstract lexical-conceptual structure that justify the surface construction (taking into account the relevant set of constraints for the given language). Here, in order to get a tractable model, we limit ourselves to the acquisition of sub-categorization frames from corpora. The task is challenging since surface forms incorporate adverbs, modifiers, interpolated clauses and some flexibility in the ordering of the arguments.  

Most approaches, including ours, are based on simple filtering techniques. If a complement appears very rarely associated with a given predicate, the acquisition process will assume that this is an incidental co-occurrence that should be left out. However, as we will see, even if this technique is efficient for high frequency items, it leaves a lot of phenomena aside. 

Following these observations, we get interested in Optimality Theory (OT). OT is based on a number of assumptions which are absolutely relevant for the lexical acquisition context \cite{kager99,mccarthy08,prince04}:
\begin{itemize}
	\item Linguistic well-formedness is relative, not absolute. Perfect satisfaction of all linguistic
constraints is attained rarely, and perhaps never.
	\item Linguistic well-formedness is a matter of comparison or competition among candidate
output forms (none of which is perfect).
	\item Linguistic constraints are ranked and violable. Higher ranking constraints can compel
violation of lower ranking constraints. Violation is minimal, however. And even low ranking
constraints can make crucial decisions about the winning output candidate.
	\item The grammar of a language is a ranking of constraints. Ranking may differ from
language to language, even if the constraints do not.
\end{itemize}

However, despite these observations, OT has been mainly applied to phonology, more rarely to morphology or syntax \cite{blache08,aarts08}. In this paper, we would like to show, on a precise example, that OT provides a very competitive framework for sub-categorization acquisition. 

In order to apply OT to lexical acquisition, we first need to model all the language properties as constraints. The task consists then in identifying the relevant set of constraints that allow one to map a lexical structure to actual (surface) constructions. Note that the task is highly challenging since constraints interact with each other, must be ranked and can be violated.

\section{From Corpus to Resources}

\subsection{OT and Syntax}

OT has been mainly applied to syntax in the framework of the Principles and Parameters (P\&P) theory developed by  Chomsky \cite{chomsky95} as part of his Minimalist Program. The central idea of P\&P is that a person's syntactic knowledge can be modeled with two formal mechanisms:
\begin{itemize}
	\item A finite set of fundamental principles that are common to all languages; e.g., a sentence must always have a subject, even if it is not overtly pronounced.
	\item A finite set of parameters that determine syntactic variability amongst languages; e.g., a binary parameter that determines whether or not the subject of a sentence must be overtly pronounced.
\end{itemize}

Within this framework, the goal of linguistics is to identify all  the principles and parameters that are universal to human languages (i.e. what defines the Universal Grammar). 

OT provides a nice framework to implement P\&P since the formalism is constraint-based. The input is a set of (universal) abstract candidate forms\footnote{This point, which is much controversial, is based on the assumption that linguistic principles---in P\&P Theory---are supposed to be universal. There is a huge literature on this hypothesis that we will not address in this paper. We do not claim any universal feature in this work; we just use OT as an interesting framework for modeling the constraints used. }. Thus, principles and parameters just have to be translated into constraints (CON); then an evaluation function (EVAL) computes  the best output given the input and the set of constraints (the principles and parameters) for a given language. 

To summarize, here are the three main components of OT: GEN (+input), CON and EVAL.

\begin{itemize}
	\item GEN takes a series of surface forms and generates an infinite number of candidates, or possible realizations of that input. A language's grammar (its ranking of constraints) determines which of the infinite candidates will be assessed as optimal by EVAL.
	\item CON includes the set of constraints to be used to determine which of the input candidates is the most likely to be accepted. 
	\item EVAL determines the best analysis among input candidates, taking into account the set of constraints CON. Given two candidates, A and B, A is better than B on a constraint hierarchy if A incurs fewer violations than B. Candidate A is better than B on an entire constraint hierarchy if A incurs fewer violations of the highest-ranked constraint distinguishing A and B. A is optimal in its candidate set if it is better on the constraint hierarchy than all other candidates. 
\end{itemize}

However, the task here is slightly different (converse) since we try to find the best underlying representation from the output (a given utterance), more precisely, we try to learn syntactic frames from data.

\subsection{Learning Syntactic Frames from Raw Data}

As already said, comprehensive and accurate lexical resources are key components of Natural Language
Processing (NLP) systems. 
Hand-crafting lexical resources
is difficult and extremely labour-intensive--- particularly as
NLP systems require statistical information about the behavior
of lexical items in context, and this statistical information changes
from one  domain to the other. For this reason automatic acquisition
of lexical resources from corpora has become increasingly popular.

One of the most useful lexical information for NLP is that related
to the predicate-argument structure. The sub-categorization frames (SCFs)
of a predicate capture the different combinations
of arguments that a given  predicate can take. For example, in French,
the verb ``\textit{acheter}'' (\textit{to buy}) sub-categorizes for a subject,
 a direct object and an indirect object (a prepositional phrase governed
by the preposition \textit{``\`a''}). This can be formalized as follows: \textit{N0 acheter N1 \`a N2}. 

Sub-categorization lexicons can benefit many NLP applications.
For example, they can be used to enhance tasks such as parsing
\cite{carroll98,arun05} and semantic classification
\cite{schulte02acl} as well as applications such as information
extraction \cite{surdeanu03} and machine translation. 
They also make it possible to infer large multilingual semantic classifications 
\cite{Sun2010}. 

Several sub-categorization lexicons are available for many languages,
but most of them have been built manually. For French these include
the large French dictionary \emph{``Le Lexique Grammaire''}
\cite{gross75} and the more recent \emph{Lefff} \cite{sagot06} and
\emph{Dicovalence} (\url{http://bach.arts.kuleuven.be/dicovalence/})
lexicons.

Some work has been conducted on automatic sub-categorization acquisition,
mostly on English \cite{brent93,manning93,briscoe97,korhonen06} but 
also on other languages, from which German is just one example \cite{schulte02lrec}.
This work has shown that although automatically built lexicons are not as accurate
and detailed as manually built ones, they can be useful for real-world tasks.
This is mostly because they provide what manually built resources
do not generally provide: statistical information about the
likelihood of SCFs for individual verbs.

In what follows, we show that statistical information, in order to yield accurate results, must take into consideration a huge number of constraints. First experiments have given interesting results but the nature and the structure of constraints must be further explored in order to strengthen the existing results. We show that OT provides an interesting framework to identify and structure the set of relevant constraints. 

\subsection{Introducing Gradience in Lexical Acquisition}

As for most linguistic questions, there is no well-established definition of what to include in a SCF, but everybody agrees that a SCF should minimally include the number and the type of the complements depending on the verb (or more generally on the predicative item considered, since adjectives and nouns can also have a SCF). Most authors agree on the fact that complements should be divided between arguments and adjuncts but the distinction between these two categories is far from obvious. Some linguistic tests exist (can the complement be deleted without changing the meaning of the sentence? Can it be moved easily? Can it be pronominalized? etc.) but none of these tests is sufficient or discriminatory enough. 

As outlined by Manning \cite{MANN2003} ``rather than maintaining a categorical argument~/ adjunct distinction and having to make in/out decisions about such cases, we might instead try to represent SCF information as a probability distribution over argument frames, with different verbal dependents expected to occur with a verb with a certain probability''. For example, from the analysis of a large news corpus, one can observe that the French verb \emph{venir (to come)} accepts the frame \emph{PP[de (from)]} with a relative frequency of 59.1\% whereas it accepts the frame \emph{PP[\`a (to)]} with a relative frequency of 5\%. This phenomenon can be seen as a kind of selectional ``preference'' of certain verbs for certain SCFs; the link with more semantic information remains to be done.

It is well known that the evaluation of probability distributions is difficult, since it is by definition dependent on a given corpus. Hand-crafted dictionaries generally do not include any frequency information. 
Moreover, very few lexical acquisition frameworks currently integrate an efficient way to deal with various phenomena such as multiword expressions (especially light verb constructions and semi-idiomatic expressions), complement optionality, etc. Therefore, current approaches have a tendency to produce two many SCFs for a given items (semi-idiomatic expressions should be recognized as such and should not be added as new SCFs associated with head verbs, optionality should be handled to reduce the number of partial SCFs). 

In the next section, we briefly present a state-of-the art system for French and its limitations; we show that the acquisition model corresponds to OT but does not take into consideration a precise enough set of constraints. We then make some proposals in order to get better results using a finer grain model of constraints.

\section{ASSCI, A State-of-the Art Subcategorization Acquisition System for French}
\label{assci}

A system for the automatic acquisition of sub-categorization frames has recently been implemented for French. 
This system called ASSCI is capable of acquiring large scale lexicons from un-annotated
corpora \cite{messiant08b,messiant-lrec08}. 

This system is close to other systems developed for example for English \cite{briscoe97,preiss07} in that it extracts SCFs from data parsed using a shallow dependency parser \cite{bourigault05} and is capable of identifying a large number of SCFs. However, unlike most other systems that  accept raw corpus data as input, it does not assume a list of predefined SCFs. The system is based on the assumption that the most relevant SCF corresponding to a given surface form will directly emerge from the application of the constraints on the various candidates, as postulated by OT.

\emph{ASSCI} takes raw corpus data as input. Input text is first tagged and syntactically analyzed.
Then, the system generates a list of candidate SCFs for each verb that occurs frequently enough in
data (in the default setting, 200 occurrences of a given verb are necessary).
\emph{ASSCI} consists of three modules: a pattern extractor
which extracts patterns for each target verb; a SCF builder which builds a list of candidate
SCFs per verb (GEN), and a SCF filter (EVAL) which filters out SCFs deemed incorrect according to predefined parameters (CON). They are described 
briefly in the following sections. For a more detailed description of \emph{ASSCI}, see \cite{messiant08b}.

\subsection{Preprocessing : Morphosyntactic Tagging and Syntactic Analysis}

The system first tags and lemmatizes corpus data using \emph{TreeTagger}
and then parses it thanks to \emph{Syntex} \cite{bourigault05}.
\emph{Syntex} is a shallow parser for French. It uses a combination of
heuristics and statistics to find dependency relations between tokens
in a sentence. It is a relatively accurate parser, e.g.~it obtained the
best precision and F-measure for written French text in the first EASY
evaluation campaign (2006).

The below example illustrates the dependency relations detected
by \emph{Syntex} \texttt{(2)} for the input sentence in
\texttt{(1)}:\\

\begin{small}
\texttt{(1) La s\'echeresse s' abattit sur le Sahel en 1972-1973 .\\
 (The drought came down on Sahel in 1972-1973.)}\\
\end{small}

\begin{scriptsize}
\texttt{(2) DetFS|le|La|1|DET;2|\\
NomFS|s\'echeresse|s\'echeresse|2|SUJ;4|DET;1 \\
Pro|se|s'|3|REF;4| \\
VCONJS|abattre|abattit|4|SUJ;2,REF;3,PREP;5,PREP;8 \\
Prep|sur|sur|5|PREP;4|NOMPREP;7 \\
DetMS|le|le|6|DET;7| \\
NomMS|sahel|Sahel|7|NOMPREP;5|DET;6 \\
Prep|en|en|8|PREP;4|NOMPREP;9 \\
NomXXDate|1972-1973|1972-1973|9|NOMPREP;8| \\
Typo|.|.|10||}
\end{scriptsize}

\bigskip

\emph{Syntex} does not make a distinction between arguments and adjuncts - rather,
each dependency of a verb is attached to the verb.

\subsection{Producing the Input (the Pattern Extractor)}

The pattern extractor collects the dependencies found by the parser for each
occurrence of a target verb.
Some cases receive special treatment in this module.
For example, if the  pronoun \textit{``se''} is one of the dependencies of a verb,
the system considers this verb like a new one.
In \texttt{(1)}, the pattern will correspond to \textit{``s'abattre''} and not to \textit{``abattre''}.
If a preposition is the head of one of the dependencies, the module explores the
syntactic analysis to find if it is followed by a noun phrase (\texttt{+SN]})
or an infinitive verb (\texttt{+SINF]}).
\texttt{(3)} shows the output of the
pattern extractor for the input in \texttt{(1)}.\\

\begin{small}
\texttt{(3) VCONJS|s'abattre~:}
\end{small}

\begin{small}
\texttt{Prep+SN|sur|PREP\_\_Prep+SN|en|PREP} 
\end{small}

\subsection{GEN (the SCF Builder)}

The SCF builder extracts SCF candidates for each verb from the output of the pattern
extractor and calculates the number of corpus occurrences for each SCF and verb
combination. The syntactic constituents used for building the SCFs are the
following:

\begin{enumerate}
 \item \texttt{SN} for nominal phrases;
 \item \texttt{SINF} for infinitive clauses;
 \item \texttt{SP[}\emph{prep}\texttt{+SN]} for prepositional phrases where the preposition is followed by a noun phrase. \emph{prep} is the head preposition;
 \item \texttt{SP[}\emph{prep}\texttt{+SINF]} for prepositional phrases where the preposition is followed by an infinitive verb. \emph{prep} is the head preposition;
 \item \texttt{SA} for adjectival phrases;
 \item \texttt{COMPL} for subordinate clauses.\\ 
 \end{enumerate}

When a verb has no dependency, its SCF is considered as \texttt{INTRANS}.

\texttt{(4)} shows the output of the SCF builder for \texttt{(1)}.\\

\begin{small}
\texttt{(4) S'ABATTRE+s'abattre ;;; SP[sur+SN]\_SP[en+SN]}
\end{small}

\subsection{CON and EVAL (SCF Filter)}
\label{filter}

Each step of the process is fully automatic, so the output of the SCF builder is noisy
due to tagging, parsing or other processing errors. It is also noisy because
of the difficulty of the argument-adjunct distinction. The latter is difficult
even for humans. 

Many criteria that have been defined are not usable in our case because
they either depend on lexical information which the parser cannot make use of
(since the task is to acquire this information) or on semantic information which even the
best parsers cannot yet learn reliably. The approach here is based on the assumption
that true arguments tend to occur in argument positions
more frequently than adjuncts. Thus many frequent SCFs in the system output
are correct.

The strategy is then to filter low frequency entries from the SCF builder output.
This is done using the maximum likelihood estimates \cite{korhonen00}. This
simple method involves calculating the relative frequency
of each SCF (for a verb) and comparing it to an empirically determined threshold.
The relative frequency of the SCF \emph{i} with the verb \emph{j} is calculated as follows:

\begin{center}
 \begin{math}
    rel\_freq(scf_{i},verb_{j}) = \dfrac{|scf_{i},verb_{j}|}{|verb_{j}|}
 \end{math}
\end{center}

\noindent
\begin{math}|scf_{i},verb_{j}|\end{math} is the number of occurrences of the SCF \emph{i} with
the verb \emph{j} and \begin{math}|verb_{j}|\end{math} is the total number of occurrences of
the verb \emph{j} in the corpus.

If, for example, the frequency of the SCF \texttt{SP[sur+SN]\_SP[en+SN]} is below the
empirically defined threshold, the SCF is rejected by the filter. The MLE filter is not
perfect because it is based on rejecting low frequency SCFs. Although relatively more low than
high frequency SCFs are incorrect, sometimes rejected frames are correct.
The filter incorporates special heuristics for cases where this
assumption tends to generate too many errors. With prepositional SCFs involving one PP or more,
the filter determines which one is the less frequent PP.
It then re-assigns the associated frequency to the same SCF without this PP.

For example, \texttt{SP[sur+SN]\_SP[en+SN]} could be split to 2 SCFs : \texttt{SP[sur+SN]}
and \texttt{SP[en+SN]}. In this  example, \texttt{SP[en+SN]} is the less frequent prepositional
phrase and the final SCF for the sentence \texttt{(1)} is \texttt{(5)}.\\

\texttt{(5) SP[sur+SN]}

\smallskip

Note that \texttt{SP[en+SN]} is here an adjunct.

\section{Some Limitations of this Approach}

This approach is very efficient to deal with large corpora. However, some issues remain. As the approach is based on automatic tools (especially parsers) that are far from perfect, the obtained resources always contain errors and have to be manually validated. Moreover, the system needs to get enough examples to be able to infer relevant information. Therefore, there is generally a lack of information for a lot of low productivity items (the famous ``sparsity problem''). 

More fundamentally, some constructions are difficult to acquire and characterize automatically. On the one hand, idioms are not recognized as such by most acquisition systems. On the other hand, some adjuncts appear frequently with certain verbs (eg. some verbs like \textit{dormir} -- \textit{to sleep }-- frequently appear with location complements). The system then assumes that these are arguments, whereas linguistic theory would say without any doubt that these are adjuncts. Lastly, surface cues are sometimes insufficient to recognize ambiguous constructions (cf. \textit{...manger une glace \`a la vanille...} vs \textit{...manger une glace \`a la terrasse d'un caf\'e...} --- \textit{to eat a vanilla ice-cream} vs \textit{to eat an ice-cream at an outdoor cafe}).

In a traditional architecture, the filtering process incorporates in one modules the set of constraints (CON) and the evaluation function (EVAL). This makes the system less readable than if the constraints were modeled apart from the EVAL function. There is thus a need to refine  the set of constraints

\section{A Solution: Provide an Explicit Modeling of the Set of Constraints (CON)}

We have shown in the previous section that a part of the errors produced were due to an over-simplification of the initial model. It is thus necessary  to take other parameters into considerations in order to yield better results. This can be done by refining the set of constraints (CON).

\subsection{Refining CON }

The issues we have reported in the previous section do not mean that automatic methods are flawed, but they have a number of drawbacks that should be addressed. The acquisition process, based on an analysis of co-occurrences of the verb with its immediate complements (along with filtering techniques) makes the approach highly functional. It is a good approximation of the problem. However, this model does not take into account external constraints. 

The analysis of the co-occurrences of the verb with its complement is meaningful but is not sufficient to fully grasp the problem. The fact that some phrasal complements (with a specific head noun) frequently co-occur with a given verb is most of the time useful, especially to identify idioms \cite{fabre08}, colligations \cite{firth57descriptive} and light verb constructions \cite{butt03}. On the other hand, the fact that a given prepositional phrase appear with a large number of verbs may indicate that the preposition introduces an adjunct rather than an argument. 

So, instead of simply capturing the co-occurrences of a verb with its complements, a number of important features should be taken into account: 

\begin{itemize}
	\item indicator of the dispersion of the prepositional phrases (PP) depending on the nature of the preposition (if a PP with a given preposition appears with a wide range of different verbs, it is more likely to be a modifier);
	\item indicator of the co-occurrence of the PP depending on the nature of the head noun (if a verb appears frequently with the same PP frame, it is more likely to form a semi-idiomatic expression);
	\item indicator of the complexity of the sentence to be processed (if a sentence is complex, its analysis is less reliable).
\end{itemize}
 
In order to do this, the pattern extractor has to be modified in order to keep most of the information that were previously rejected as not relevant. These indicators then need to be calculated so as to be taken into account by EVAL.

\subsection{Modifying EVAL}

All the constraints can be evaluated separately, so as to obtain for each of them an ideal evaluation of the parameter. There are two ways of doing this: \textit{i}) by automatically inferring the different weights from a set of annotated data or \textit{ii}) by estimating the results of various manually defined weights. We are currently using this last method since data annotation is very costly. However, the first approach would certainly lead to more accurate results. 

The weight and the ranking of the different constraints must then be examined. A linear model can provide a first approximation but there are surely better ways to integrate the different constraints. Some studies provide some cues but they need to be proper evaluated in order to be integrated in this framework \cite{blache08}.

\subsection{Manual Validation}

Lastly, the approach requires a manual validation. Rather than leaving the validation process apart for further examination by a linguist, we propose to integrate it in the acquisition process itself. Taking into consideration the number of examples and the complexity of the sentences used for training, it is possible to associate confidence scores with the different constructions of a given verb: the linguist is then able to quickly focus on the most problematic cases. It is also possible to propose tentative constructions to the linguist, when not enough occurrences are available for training. In the end, when too few examples are available, the linguist can provide relevant information to the machine. However, with a well-designed and dynamic validation process, it is possible to obtain accurate and comprehensive lexicons, using only a small fraction of the time that would be necessary to manually develop  a lexicon from scratch.

\section{Conclusion}

Tn this paper, we have proposed a new approach for the automatic acquisition of lexical knowledge from corpora using Optimality Theory. 
Using this model, it is possible to represent a large part of the language activity through constraints.
We have shown that the individual evaluation of each constraint yields very accurate and precise results. 

An implementation of this model is currently being done for Japanese \cite{marchal2012}. The model provides a better integration of the linguistic contraints within the automatic processing system. First results were competitive with other approaches while providing a more accurate linguistic description.

\bibliographystyle{splncs}
\bibliography{cicling2014}

\end{document}